\documentclass[10pt,twocolumn,letterpaper]{article}

\usepackage{cvpr}              % To produce the CAMERA-READY version
\usepackage{graphicx}
\usepackage{amsmath}
\usepackage{amssymb}
\usepackage{xcolor}
\usepackage{multirow}
\usepackage{tabularx} % Package for tabularx environment
\usepackage{booktabs}
\usepackage{algorithm}
\usepackage{float}
\usepackage{cuted}
\usepackage[noend]{algpseudocode}

\usepackage[pagebackref,breaklinks,colorlinks]{hyperref}

\usepackage[capitalize]{cleveref}
\crefname{section}{Sec.}{Secs.}
\Crefname{section}{Section}{Sections}
\Crefname{table}{Table}{Tables}
\crefname{table}{Tab.}{Tabs.}

\begin{document}

%%%%%%%%% TITLE - PLEASE UPDATE

% \title{Vision Transformers and Motion Modality Fusion for Enhanced Deepfake Detection}

% \title{Unmasking Deception: Masked Autoencoding Spatiotemporal Transformers for Enhanced Video Deepfake Detection} 

% \title{Deception Unmasked: Masked Autoencoding Spatiotemporal Transformers for Enhanced Deepfake Detection} 

\title{Unmasking Deepfakes: Masked Autoencoding Spatiotemporal Transformers for Enhanced Video Forgery Detection}

 % Masked Autoencoding Spatiotemporal Deepfake Transformers

% \title{Facial Forgery Detection with Multi-domain Transfer Learning}

% \title{MASDT:  Vision Transformer with Motion Modality for Accurate Detection of Deepfakes}

% \title{MASDT: Fusing Optical Flow and Masked Autoencoders for Robust Facial Forgery Detection on Limited Data}

\author{Sayantan Das$^{1}$, Mojtaba Kolahdouzi$^{1}$, Levent Özparlak$^{2}$, Will Hickie$^{2}$, Ali Etemad$^{1}$ \\
$^{1}$Queen's University, Canada\\
$^{2}$Irdeto BV\\
% {\tt\small firstauthor@i1.org}
% <will.hickie@irdeto.com>
%  <levent.ozparlak@irdeto.com>
% For a paper whose authors are all at the same institution,
% omit the following lines up until the closing ``}''.
% Additional authors and addresses can be added with ``\and'',
% just like the second author.
% To save space, use either the email address or home page, not both
% \and
% Second Author\\
% Institution2\\
% First line of institution2 address\\
% {\tt\small secondauthor@i2.org}
}
\maketitle

%%%%%%%%% ABSTRACT
\begin{abstract}
    We present a novel approach for the detection of deepfake videos using a pair of vision transformers pre-trained by a self-supervised masked autoencoding setup. Our method consists of two distinct components, one of which focuses on learning spatial information from individual RGB frames of the video, while the other learns temporal consistency information from optical flow fields generated from consecutive frames. Unlike most approaches where pre-training is performed on a generic large corpus of images, we show that by pre-training on smaller face-related datasets, namely Celeb-A (for the spatial learning component) and YouTube Faces (for the temporal learning component), strong results can be obtained. We perform various experiments to evaluate the performance of our method on commonly used datasets namely FaceForensics++ (Low Quality and High Quality, along with a new highly compressed version named Very Low Quality) and Celeb-DFv2 datasets. Our experiments show that our method sets a new state-of-the-art on FaceForensics++ (LQ, HQ, and VLQ), and obtains competitive results on Celeb-DFv2. Moreover, our method outperforms other methods in the area in a cross-dataset setup where we fine-tune our model on FaceForensics++ and test on CelebDFv2, pointing to its strong cross-dataset generalization ability. We make the code publicly available at \href{http://ucalyptus.github.io/masdt}{http://ucalyptus.github.io/masdt}.
\end{abstract}

%%%%%%%%% BODY TEXT
\section{Introduction}

\begin{figure}[t]
 \centering
 \includegraphics[width=\columnwidth]{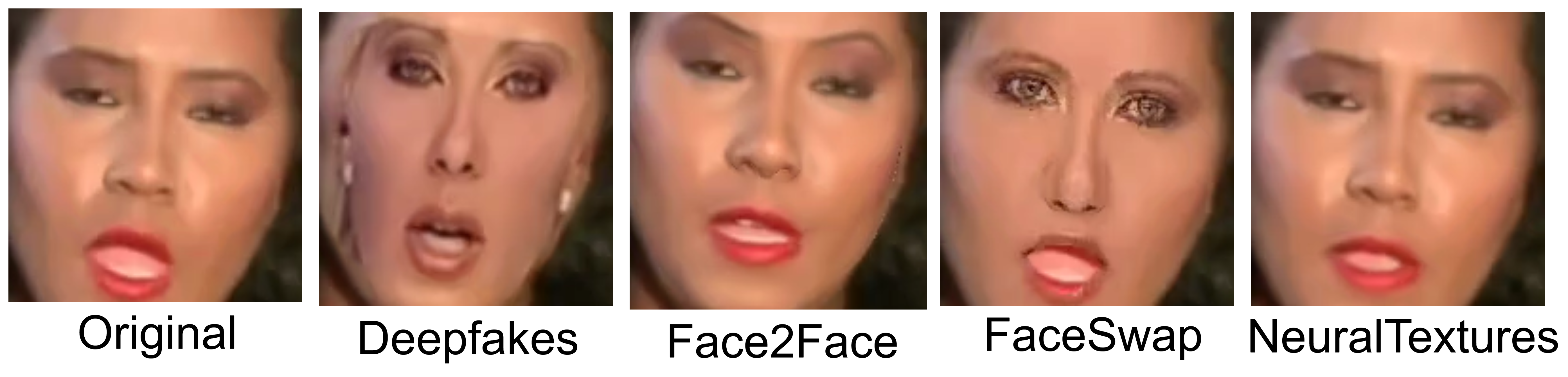}
 \caption{A facial image along with the four manipulated versions from the FaceForensics++ dataset.}
 \label{fig:figure-1}
 \end{figure}

Facial forgery detection, also known as deepfake detection, is a rapidly growing field with important real-world applications \cite{tiwari2023leveraging}. With the recent explosion in the success of sophisticated deep generative models \cite{kingma2014semi,xu2015overview,pmlr-v32-rezende14}, it has become increasingly easy to generate highly realistic fake images and videos (see Figure \ref{fig:figure-1} for an example of a real image along with four manipulated versions). The advancements in artificial intelligence have made it possible to create deepfakes that are nearly indistinguishable from genuine content, making the detection process even more challenging. This has led to a growing concern about the potential for malicious actors to use these tools for nefarious purposes, such as spreading disinformation, manipulating public opinion, or even causing social unrest. Given the potential risks associated with deepfakes, the importance of developing effective detection methods cannot be overstated \cite{10.1145/3503250,radford2021learning,ldm}.

The field of deepfake detection has seen considerable progress in recent years with a number of sophisticated techniques being proposed in the area \cite{iet2021survey,rana2022deepfake}. However, in the context of detecting manipulated content in videos, many existing methods primarily focus on spatial features extracted from individual frames \cite{cao2022end}. This approach can lead to the overlooking of temporal dynamics that evolve throughout video sequences. This strategy can result in limitations, as temporal artifacts such as flickering and motion discontinuities, are common indicators of deepfake manipulation. Furthermore, sophisticated deepfakes may exhibit subtle spatial inconsistencies that manifest over time, necessitating an integrated analysis of both spatial and temporal information. Moreover, we hypothesize that capturing subtle \textit{spatiotemporal} inconsistencies that are often caused by different deepfake generation methods, could significantly enhance performance by learning representations that generalize to unseen forgery methods, which is often a challenging problem in this area.

In response to the challenges mentioned above, in this paper, we present a novel approach to deepfake detection that consists of two distinct components. One component learns spatial information from individual RGB frames of the videos, while the second component leverages optical flow fields to learn temporal consistency across the video. Both components utilize vision transformers \cite{dosovitskiy2020vit}, which we train in two steps. First, inspired by \cite{he2021masked} we pre-train the models in an autoencoding setup using a self-supervised reconstruction scheme. Second, we discard the reconstruction decoder and add a new classification head to each encoder, where they are fine-tuned for deepfake detection, followed by score-level fusion of the results. We pre-train the spatial learning and temporal consistency learning encoders with CelebFaces-Attributes (Celeb-A) \cite{liu2015faceattributes} and YouTube Faces \cite{wolf2011face} datasets respectively. For downstream deepfake detection, we evaluate our approach on the FaceForensics++ (High Quality) and FaceForensics++ (Low Quality) datasets \cite{roessler2019faceforensicspp} which employ compression factors of 23\% and 40\%, respectively, in addition to the CelebDFv2 dataset \cite{li2020celebdf}. Additionally, we synthesize a more challenging variation of the FaceForensics++ dataset, which we call FaceForensics++ (Very Low Quality) by compressing the data with a rate of 65\%. This dataset is then used to further evaluate our approach against existing techniques in the presence of extreme compression artifacts. Experimental results demonstrate that our method achieves state-of-the-art performance on FaceForensics++ (LQ and HQ) datasets and highlight the efficacy of our approach in detecting deepfakes across diverse compression levels. Ablation studies demonstrate the importance of different components of our method. Lastly, state-of-the-art results when fine-tuning our model on FaceForensics++ and testing it on CelebDFv2 (cross-dataset evaluation) demonstrates the strong generalization of our method.

Our contributions in this paper are summarized as follows: (\textbf{1}) We propose a new approach for effective facial forgery detection. Our method uses a vision autoencoding transformer and is pre-trained in a self-supervised masked reconstruction setup. Our solution consists of two main components which learn spatial (RGB) and temporal consistency information (optical flow fields) separately. (\textbf{2}) We leverage relatively small datasets, namely Celeb-A and YouTube Faces, for pretraining our transformers, and achieve state-of-the-art results in the downstream task of deepfake detection on FaceForensics++ (LQ, HQ, and VLQ) datasets and competitive results on Celeb-DFv2. 
(\textbf{3}) We make our code available online at \href{http://ucalyptus.github.io/masdt}{http://ucalyptus.github.io/masdt} to contribute to the area and enable reproducibility.

\section{Related Work} 

% \subsection{Deepfake Detection}
Deepfake detection has traditionally been addressed as a binary classification task \cite{rana2022deepfake}, where the objective is to discern between authentic and manipulated media. The application of deep learning models, particularly convolutional neural networks (CNNs), has been central to achieving this objective \cite{Chollet2017XceptionDL,Cozzolino2017RecastingRL,Afchar2018MesoNetAC}. Authors of FaceForensics++ dataset \cite{roessler2019faceforensicspp} used Xception network, which was one of the best-performing architectures at the time, to perform deepfake detection via transfer learning \cite{Chollet2017XceptionDL}.

Researchers proposed a method in \cite{Cozzolino2017RecastingRL} that utilizes residual-based descriptors in the form of a constrained CNN for image forgery detection. This approach aims to capture and analyze the residual noise present in manipulated images, which can be a strong indicator of forgery. In contrast, another method introduced a deep learning approach that focuses on the mesoscopic properties of images \cite{Afchar2018MesoNetAC}. In this context, 
% of image analysis and deep learning, 
`mesoscopic' refers to properties or features that fall between the small scale (microscopic) and the large scale (macroscopic). By concentrating on mesoscopic features, the model can capture subtle artifacts and inconsistencies in manipulated images, potentially making it more effective in detecting forgeries.

% Attention mechanisms, as introduced in \cite{vaswani2017attention}, have been combined with CNNs in various works \cite{zhao2021multi,sun2022information} to enhance interpretability and facilitate the identification of manipulated regions. These attention-based models generate attention maps that highlight regions contributing significantly to the detection decision. For example, the study in \cite{zhao2021multi} utilized attention maps generated by deep semantic features to outline crucial regions that contributed to the classification result. These attention maps guide the aggregation of low-level textural features and high-level semantic features, which helps to capture more subtle artifacts in the image. Furthermore, a new attention mechanism was proposed in \cite{sun2022information} that calculates the self-information from the input feature map and outputs a discriminative attention map. This attention map emphasizes regions that contribute significantly to the detection decision, enhancing the model's ability to identify manipulated areas.

Various studies have utilized frequency analysis to detect inconsistencies that arise during deepfake creation  \cite{liu2021spatial,Chen_Yao_Chen_Ding_Li_Ji_2021}. In \cite{liu2021spatial}, the researchers employed the phase spectrum for forged face image detection, showing that CNNs can identify additional implicit phase spectrum features that are advantageous in detecting face forgeries. Concurrently, the study in \cite{Chen_Yao_Chen_Ding_Li_Ji_2021} developed a multi-scale patch similarity module to specifically model second-order relationships between distinct local regions, forming a similarity pattern through pairwise cosine measurements. These patterns distinguish real from forged regions by recognizing differences such as irregular textures and high-frequency noise.

Self-supervised learning (SSL) has been explored to address the issue of limited labeled data for deepfake detection \cite{chen2022self,zhao2022selfsupervised,9428368,knafo2022fakeout,xu2022supervised}. For instance, self-supervised learning was employed in \cite{chen2022self} with an auxiliary task specifically designed for deepfake detection, using a synthesizer and adversarial training framework to dynamically generate forgeries. This approach enriches diversity and strengthens sensitivity to produce strong results. In the method proposed in \cite{zhao2022selfsupervised}, mouth motion representations were learned by encouraging close-paired video and audio representations, while keeping unpaired ones diverse. The study in \cite{9428368} proposed a decoupling strategy to separate facial authenticity and compression relevance, implemented through a joint self-supervised learning approach using compression ratios as self-supervised signals. Another study utilized a multi-modal backbone trained in a self-supervised manner and adapted it to the video deepfake domain \cite{knafo2022fakeout}. These self-supervised models leverage unlabeled data to learn useful representations for detection tasks. Contrastive learning is another common pre-text learning approach often considered for deepfake detection \cite{fung2021deepfakeucl,xu2022supervised}. In the study by \cite{fung2021deepfakeucl}, two different transformed versions of a face image were generated using two distinct transformations. The agreement between these transformed images is maximized after they are passed through an encoder network and a projection head network, effectively training the model without supervision signals. On the other hand, another study employed supervised contrastive learning to learn common features between instances of the same class, while distinguishing between samples from different classes \cite{xu2022supervised}.

In recent studies, researchers have sought to combine multiple modalities, such as visual, audio, and temporal information, to improve detection performance in deepfake detection tasks \cite{liu2021spatial,cai2022really,ilyas2023avfakenet}. These multi-modal approaches provide a comprehensive view of the media, making them more robust against limitations specific to individual modalities \cite{khalid2021fakeavceleb}. While frequency-based modalities have been employed in multi-modal deepfake detection solutions \cite{liu2021spatial}, we believe that optical flow has the potential to serve as an effective alternative source of information. 
% Furthermore, we aim to address generalizability and robustness in representation learning by employing the MAE framework based on vision transformers. 
Prior works such as \cite{Amerini2019DeepfakeVD,CALDELLI202131,9892905,10.1145/3579731.3579810} have used optical flow features followed by a classifier for deepfake detection. 
% Jiang et al \cite{10.1145/3579731.3579810} leverages attention-fusion mechanisms using optical flow information. 
In contrast to these, our method, for the first time, leverages optical flow information alongside RGB features in a masked autoencoding setup to improve cross-dataset generalizability and robustness. To our knowledge, prior deepfake detection methods involving ViTs \cite{zhuang2022uiavit,hu2023mover,shi2023real,Heo2023,wodajo2021deepfake} have not used optical flow information to capture the temporal inconsistencies. 
% Furthermore, we aim to address generalizability and robustness in representation learning by employing the MAE framework based on vision transformers. 

\begin{figure*}[t]
    \centering
    \begin{centering}        \includegraphics[height=0.55\linewidth]{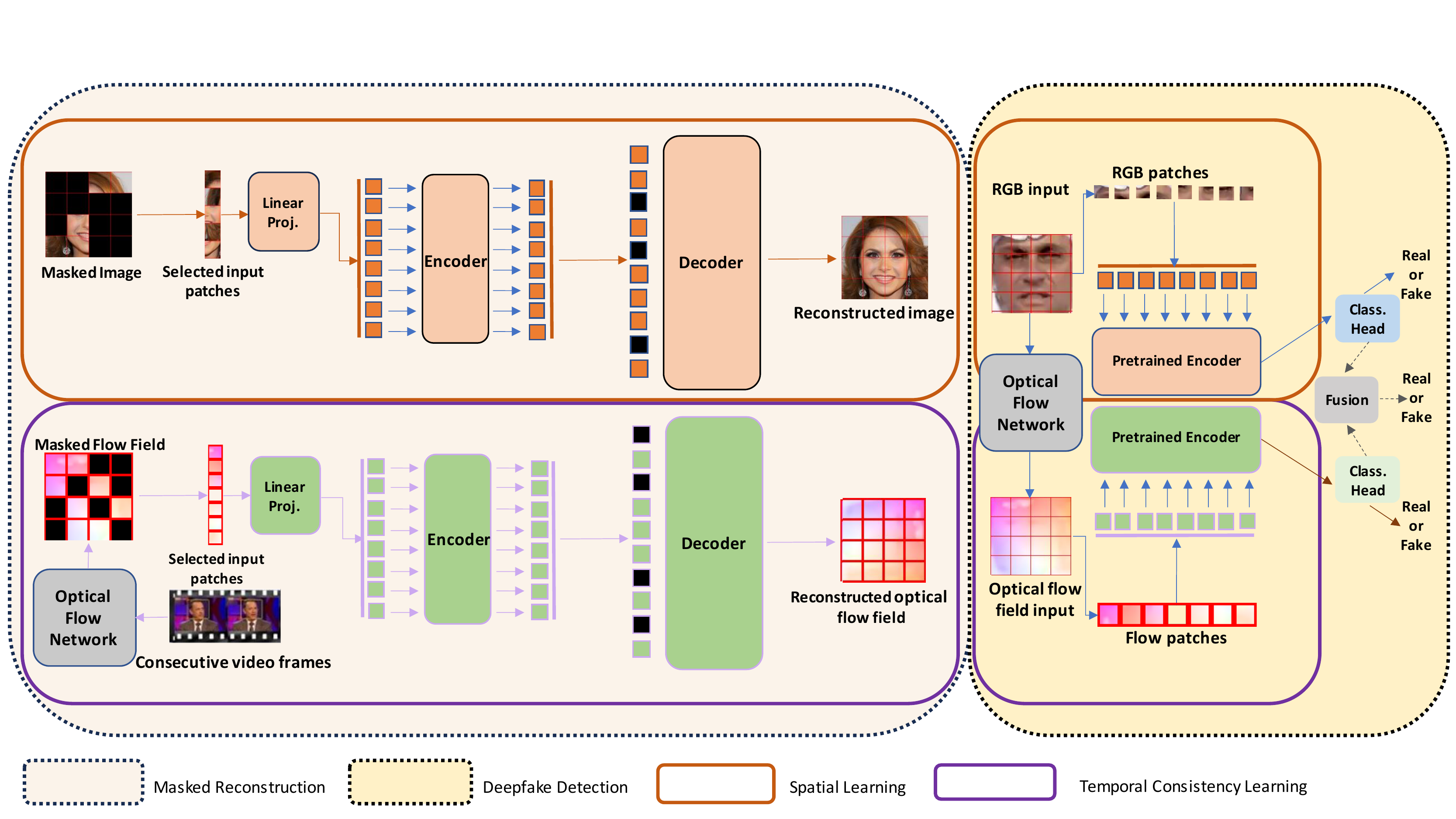}
    \end{centering}
    \caption{An overview of our method (MASDT), which includes the masked facial reconstruction and deepfake detection steps.}
    \label{fig:modelarch}
\end{figure*}

\section{Proposed Methodology}

\subsection{Overview}\label{strategy}

Our proposed approach titled Masked Autoencoding Spatiotemporal Deepfake Transformer (MASDT) consists of two components: spatial learning and temporal consistency learning. The spatial learning component has the objective of learning robust spatial features from the RGB images, while temporal consistency learning aims to extract temporal features from optical flow fields derived from the input images. We fuse the classification outputs derived from the spatial and temporal consistency learning components. Both these components follow a self-supervised autoencoding approach in a \textit{two-step} process.

The first step involves a self-supervised pre-training strategy which involves both of the MASDT components in a data reconstruction task. We discuss this strategy in section \ref{RE-RD}. The second step is the downstream task of deepfake detection,
wherein we re-purpose components trained in the previous step to perform the classification of deepfake data through a model fine-tuning process, followed by fusion of information from both components (spatial learning and temporal consistency learning). This is discussed in detail in Section \ref{CE}. Before we discuss each of the two steps, we discuss the optical flow field generation strategy in Section \ref{OF}. A general scheme of the MASDT strategy is presented in Figure \ref{fig:modelarch}.

\subsubsection{Optical flow field estimation}\label{OF}
We utilize a CNN model named PWC-Net for generating optical flow fields \cite{Sun2018PWC-Net}. Let the model for estimating optical flow be $F_{\theta}$, and two consecutive frames be $f_t$ and $f_{t+1}$. Accordingly, the estimated optical flow $\Phi_t$ can be denoted by:
\begin{equation}
\Phi_t = F_\theta(f_t, f_{t+1}),
\end{equation}
where $\Phi_t$ is a 3-channel optical flow matrix of size $H \times W \times 3$ representing the flow field between the consecutive frames.

\subsection{Masked Facial Reconstruction}\label{RE-RD}
The first step of our approach utilizes a masked self-supervised auto-encoder which learns to reconstruct original facial images, given partial observations \cite{he2021masked}. This auto-encoder reconstruction pipeline consists of two blocks: a reconstruction encoder, which captures a latent representation from the visible portions of each image, and a reconstruction decoder that aims to reconstruct the masked sections of the image using this latent representation. In this procedure, the encoder is trained to extract robust spatial features from masked facial images, eliminating noise and redundancy while transforming the reconstruction task into a challenging process that requires generalizing features to represent a small subset of available data \cite{ibanez2022masked}. Consequently, by masking portions of the facial image using random spatial pixels or patches, we can avoid a potential location bias toward image reconstruction, which can be critical for the detection of deepfake images.

The goal of the decoder is to use the features obtained from the latent space by the encoder to reconstruct the masked information from the original facial image. We train this reconstruction encoder-decoder pair using a simple mean squared error (MSE) reconstruction loss $\mathcal{L}_r$:
\begin{equation}
\mathcal{L}_r = \frac{1}{N}\sum_{i=1}^{N}(y_i - \hat{y}_i)^2 ,
\end{equation}
where $N$ represents the number of sampled patches, and $\hat{y}_i$ and $y_i$ are the $i$th output and expected $i$th output, respectively.

We perform the above masked reconstruction task for both the components independently where we employ the encoder-decoder pairs for reconstructing RGB images $y$ and optical flow fields $\Phi$ for spatial learning and temporal consistency learning respectively, which is also referred to as pre-training in self-supervised learning literature. This prepares the encoders for the fine-tuning step mentioned in the next section.

\subsection{Deepfake Detection}\label{CE}
The second step of MASDT is aimed at the supervised training for the classification of deepfake images. For this purpose, we employ the encoders that learned to extract robust representations in the reconstruction pipeline. Thus, to perform binary classification, a classification head consisting of a simple MLP is attached to each of the pre-trained encoders.

We adopt a dual-encoder setup for the fine-tuning process, utilizing the spatial learning encoder $\theta_s$ and the temporal consistency learning encoder $\theta_t$. These encoders were previously trained in the initial step of our proposed solution. In the process of fine-tuning for a binary classification task, we employ a binary cross-entropy loss, denoted as $\mathcal{L}_b$. The formula for this loss is as follows:
\begin{equation}
\mathcal{L}_b = -\frac{1}{M}\sum_{j=1}^{M} \left[ o_j \log(\hat{o}_j) + (1 - o_j) \log(1 - \hat{o}_j) \right],
\end{equation}
Here, $\hat{o}_j$ is the predicted output from the network, $o_j$ represents the actual or target class (either 0 or 1), and $M$ denotes the total count of samples in the batch.

\subsubsection{Dual Modality Fusion}\label{DMF}
To further harness the strengths of both $\theta_s$ and $\theta_t$, we use a simple fusion mechanism. This method aims to exploit the complementary information that each encoder provides, thereby improving the overall classification performance. The fusion process begins with the individual outputs from $\theta_s$ and $\theta_t$, denoted as $\hat{o}_s$ and $\hat{o}_t$ respectively, which are then combined to create a fused score-level prediction, $\hat{o}_f$. Mathematically, this can be expressed as:
\begin{equation}
\hat{o}_f = \alpha \cdot \hat{o}_s + (1 - \alpha) \cdot \hat{o}_t,
\end{equation}
where $\alpha$ is a fusion weight that determines the contribution of each encoder to the final output.

\section{Experiments}
In this section, we present the specifics and details of our method and experiments, describe the datasets used, and discuss the ablation studies conducted to validate the impact of different components of our proposed solution.

\subsection{Implementation Details}
In this section, we outline the implementation details of our deepfake detection method, which incorporates both RGB and optical flow modalities. Our experiments are conducted using the PyTorch framework \cite{NEURIPS2019_9015} on 4 Nvidia A100 GPUs, each with 40 GB of vRAM. We generate optical flow fields using the PWC-Net present in the MMFlow toolbox \cite{2021mmflow}.

Our method's performance is evaluated using the top-1 accuracy, which denotes the percentage of correctly classified deepfake and real videos out of the total number of videos in the test set. This metric is widely used in deepfake detection tasks as it provides a clear indication of a model's ability to distinguish between real and fake videos. Accuracy and area under the curve (AUC) are presented as the metrics for our experiments, following other publications in the area.

For evaluation purposes, we use the FaceForensics++ (LQ and HQ) and CelebDFv2 datasets (the details of these datasets are presented in the next Section) and divide them into training, validation, and test sets, ensuring an even distribution of deepfake and real videos across all sets following the instructions provided in the original dataset papers \cite{roessler2019faceforensicspp,li2020celebdf}. Data augmentation techniques such as random cropping, horizontal flipping, color jittering, and MixUp augmentation, are employed to improve our model's robustness to input data variations. MixUp augmentation \cite{zhang2018mixup} involves generating new training samples by taking linear combinations of input data and their corresponding labels, which encourages the model to learn smooth and robust features. %We discuss the datasets used in reconstruction and deepfake detection steps in the following subsection.
In addition to MixUp, the model employs CutMix \cite{yun2019cutmix} data augmentation technique with default settings (alpha set to 0, probability set to 1, and switch probability set to 0.5). Label smoothing is applied with a smoothing factor of 0.1. A drop path rate of 0.1 is used for stochastic depth regularization.

Input images are resized to $224 \times 224$, with patches of $16 \times 16$. We observe that a masking ratio of 90\% is optimal for pre-training. We use the transformer architecture \cite{dosovitskiy2020vit} with a default Vit-B configuration as our model. The model is trained using the AdamW optimizer, with a weight decay of 0.05, a base learning rate of $5\times10^{-4}$, and layer decay of 0.8. The learning rate is scaled according to an effective batch size of 64. We train the model for 300 epochs, using a gradient accumulation of 1 iteration. We utilize a distributed training approach with distributed evaluation. The CUDA benchmark is enabled, and the model is trained on available CUDA devices. For fine-tuning, the model is initialized with our pre-trained weights from the first step (self-supervised pre-training), and position embeddings are interpolated accordingly.

\subsection{Datasets}\label{datasets}
We use the FaceForensics++ (LQ), FaceForensics++ (HQ) \cite{roessler2019faceforensicspp}, Celeb-DFv2 \cite{li2020celebdf}, Celeb-A \cite{liu2015faceattributes}, and YouTube Faces \cite{wolf2011face} datasets. The first three datasets are employed for evaluating our proposed method, while the latter two are utilized for pre-training only. Below, we provide a detailed description of each dataset:

\noindent \textbf{FaceForensics++ (LQ)} \cite{roessler2019faceforensicspp} simulates various scenarios where manipulated videos appear in compressed formats. With a 40\% compression factor using the H.264 video compression standard, the LQ version introduces artifacts that may be present in real-world cases. This dataset challenges researchers to develop techniques capable of detecting manipulations even when the video quality is degraded due to compression.

\noindent \textbf{FaceForensics++ (HQ)} \cite{roessler2019faceforensicspp} maintains a higher quality (compression factor of 23\%) compared to the LQ version, enabling researchers to study deepfakes and other manipulations with greater detail and less information loss due to compression. Both FaceForensics++ versions contain over 1000 original videos, with manipulated videos created using various methods, such as DeepFakes \cite{deepfakes_faceswap}, FaceSwap \cite{marekkowalski_faceswap}, Face2Face \cite{thies2016face2face}, and NeuralTextures \cite{Thies2019DeferredNR}. These datasets cover a wide range of manipulation methods, allowing researchers to test their detection algorithms on diverse types of deepfakes.

In order to further push our method to the limit and challenge its detection ability in the presence of significant compression artifacts, we create an even more compressed version in comparison to FaceForensics++ (LQ), which we call \textbf{FaceForensics++ (VLQ)} where VLQ stands for very low quality. To generate this variant of the dataset, we take the original non-compressed videos of FaceForensics++ and compress them by a compression factor of 65\%, which we will also use in our experiments besides the datasets with two standard compression ratios. For this purpose, we use the FFMPEG framework \cite{tomar2006converting}.

\noindent \textbf{Celeb-DFv2} \cite{li2020celebdf} includes 590 original videos collected from YouTube, featuring subjects of varying ages, ethnicities, and genders, as well as 5639 corresponding DeepFake videos. The Celeb-DF dataset's average video length is 13 seconds, and all videos have a standard 30 FPS frame rate.

\noindent \textbf{Celeb-A} (CelebFaces-Attributes) \cite{liu2015faceattributes} is a large-scale collection of over 200,000 celebrity images, with 40 attribute labels annotated for each image. The dataset comprises diverse subjects and captures various facial expressions, poses, and lighting conditions.

\noindent \textbf{YouTube Faces} \cite{wolf2011face} is a comprehensive collection of videos from YouTube focusing on individuals' faces. It contains over 3,000 annotated videos of 1,595 people, offering diverse subjects with different ethnicities, ages, and genders. Each video in the dataset is labeled with the corresponding subjects' identities, and is often used for face recognition and verification tasks. It captures various poses, expressions, illuminations, and occlusions.

\subsection{Pre-training Strategy}

For pre-training the RGB modality in our proposed method, we utilize the Celeb-A dataset instead of the typically used ImageNet \cite{deng2009imagenet}. Celeb-A is considerably smaller than ImageNet, as Celeb-A contains 200,000 images whereas ImageNet contains over 14 million images. This reduced size allows for faster pre-training and lower computational requirements, making the process more efficient and accessible to a wider range of researchers and practitioners. Celeb-A is specifically tailored for facial tasks, consisting exclusively of human face images. In contrast, ImageNet covers many object categories and may not be as well-suited and efficient for facial analysis. By pre-training our model on Celeb-A, we ensure that the initial features learned by the model are more relevant to facial structures, expressions, and attributes, which can ultimately contribute to a more effective deepfake detection system.

For pre-training the optical flow modality in our method, we utilize the YouTube Faces dataset. This dataset provides video data, essential for optical flow calculation. Naturally, datasets of images such as ImageNet and Celeb-A cannot be used for optical flow generation. Moreover, the YouTube Faces dataset is specifically designed for facial analysis tasks as it consists exclusively of human face videos. By pre-training our model on this dataset, we ensure that the initial features learned by the temporal consistency encoder can better capture information such as facial structures, expressions, and attributes, ultimately contributing to a more effective deepfake detection system.

\subsection{Results}
In this section, we present the outcome of our experiments, which assess the performance of the proposed method for deepfake detection on the FaceForensics++ and Celeb-DFv2 datasets. Our evaluation concentrates on the effectiveness of integrating both RGB and optical flow modalities, as well as the impact of pre-training on the Celeb-A and YouTube Faces datasets. By contrasting our approach with existing methods and baseline models, we aim to evaluate the benefits of our technique in accurately identifying deepfakes under a range of conditions.

In Table \ref{tab:top1} we present the top-1 accuracy and AUC scores of our proposed method compared to the current state-of-the-art approaches. The table presents the quantitative results for various deepfake detection techniques available in the FaceForensics++ dataset with both high and low quality settings. It can be observed that our proposed method achieves the highest accuracy and AUC scores in both quality settings, surpassing the prior works and setting a new state-of-the-art.

\begin{table}[t]
\centering
\small
\setlength
\tabcolsep{2pt}
\caption{Quantitative results for ACC and AUC on the FaceForensics++ dataset with both quality settings (LQ and HQ). The results are arranged in ascending order on the basis of ACC (LQ).}
\begin{tabular}{l cccc}
\hline
\multirow{2}{*}{\textbf{Methods}} & \textbf{ACC} & \textbf{AUC} & \textbf{ACC} & \textbf{AUC} \\ %\cline{2-5}
& \textbf{(HQ)} & \textbf{(HQ)} & \textbf{(LQ)} & \textbf{(LQ)} \\ \hline\hline
Steg. Features ~\cite{Fridrich2012RichMF} & 70.97\% & - & 55.98\% & - \\ %\hline 
LD-CNN ~\cite{Cozzolino2017RecastingRL} & 78.45\% & - & 58.69\% & - \\ %\hline
CP-CNN ~\cite{Rahmouni2017DistinguishingCG} & 79.08\% & - & 61.18\% & - \\ %\hline
Face X-ray ~\cite{Li2020FaceXF} & - & 87.40\% & - & 61.60\% \\ %\hline
C-Conv ~\cite{Bayar2016ADL} & 82.97\% & - & 66.84\% & - \\ %\hline
MesoNet ~\cite{Afchar2018MesoNetAC} & 83.10\% & - & 70.47\% & - \\ %\hline
FakeCatcher \cite{9141516} & 94.65\% & - & - & - \\ %\hline
Two-branch RN ~\cite{2020arXiv200803412M} & 96.43\% & 88.70\% & 86.34\% & 86.59\% \\ %\hline
Xception ~\cite{Rssler2019FaceForensicsLT} & 95.73\% & - & 86.86\% & - \\ %\hline
LipsDontLie \cite{DBLP:conf/cvpr/HaliassosVPP21} & - & 97.10\% & - & - \\ %\hline
Capsule Net \cite{Tolosana2020DeepFakesEA} & - & 99.50\% & - & - \\ %\hline
SLADD \cite{chen2022self} & - & 98.40\% & - & - \\ %\hline
MADD \cite{zhao2021multi} & 97.60\% & 99.29\% & 88.69\% & 90.40\% \\ %\hline
Self Info. Att. \cite{sun2022information} & 97.64\% & 99.35\% & 90.23\% & 93.45\% \\ %\hline
F3-Net ~\cite{Qian2020ThinkingIF} & 97.52\% & 98.10\% & 90.43\% & 93.30\% \\ %\hline
E2E Learning \cite{cao2022end}& 97.06\% & 99.32\% & 91.03\% & 95.02\% \\ %\hline
Local Relation Learning \cite{Chen_Yao_Chen_Ding_Li_Ji_2021} & 97.59\% & 99.46\% & 91.47\% & 95.21\% \\ %\hline
Ours & \textbf{98.19\%} & \textbf{99.67\%} & \textbf{97.79\%} & \textbf{98.45\%} \\ \hline
% \hline
\end{tabular}
\label{tab:top1}
\end{table}

In our experiments, we assess the performance of different deepfake generation methods in the FaceForensics++ (LQ) dataset, comprising four distinct techniques: DeepFakes (DF) \cite{deepfakes_faceswap}, Face2Face (FF) \cite{thies2016face2face}, FaceSwap (FS) \cite{marekkowalski_faceswap}, and NeuralTextures (NT) \cite{Thies2019DeferredNR}, as illustrated in Table \ref{tab:intraFF}. In this table, we present a breakdown of the performance of our method and others across these four deepfake generation methods, and compare the accuracy with other state-of-the-art approaches. The results indicate that our method achieves strong results across all four manipulation techniques, particularly in the FF and FS methods, and generates competitive results for the other two. These findings demonstrate the effectiveness of our approach in detecting manipulated face images across different forgery approaches.

\begin{table}[t]
\centering
\small
\setlength
\tabcolsep{3pt}
\caption{Quantitative results (ACC) on the FaceForensics++ (LQ) dataset with four manipulation methods: DeepFakes (DF), Face2Face (FF), FaceSwap (FS), and NeuralTextures (NT).}
\begin{tabular}{lllll}
\hline
\textbf{Methods} & \textbf{DF} \cite{deepfakes_faceswap} & \textbf{FF} \cite{thies2016face2face} & \textbf{FS} \cite{marekkowalski_faceswap} & \textbf{NT} \cite{Thies2019DeferredNR}\\
\hline\hline
Steg. Features \cite{Fridrich2012RichMF} & 67.00\% & 48.00\% & 49.00\% & 56.00\% \\
LD-CNN  \cite{Cozzolino2017RecastingRL} & 75.00\% & 56.00\% & 51.00\% & 62.00\% \\
C-Conv \cite{Bayar2016ADL} & 87.00\% & 82.00\% & 74.00\% & 74.00\% \\
CP-CNN \cite{Rahmouni2017DistinguishingCG} & 80.00\% & 62.00\% & 59.00\% & 59.00\% \\
MesoNet  \cite{Afchar2018MesoNetAC}& 90.00\% & 83.00\% & 83.00\% & 75.00\% \\
Xception \cite{Rssler2019FaceForensicsLT} & 96.01\% & 93.29\% & 94.71\% & 79.14\% \\
F3-Net \cite{Qian2020ThinkingIF} & 97.97\% & 95.32\% & 96.53\% & 83.32\% \\
Local Relation Learning  \cite{Chen_Yao_Chen_Ding_Li_Ji_2021} & \textbf{98.84} \% & 95.53\% & 97.53\% & \textbf{89.31\%} \\
Ours & 97.84\%  & \textbf{96.27\%} & \textbf{97.89\%} & 78.23\% \\
\hline
\end{tabular}
\label{tab:intraFF}
\end{table}

Next, we evaluate the performance of our method compared to other recent methods on the Celeb-DFv2 dataset and present the performance in Table \ref{tab:celdf-finetune}. It can be observed that our method achieves results competitive to the current state-of-the-art \cite{cao2022end}.

\begin{table}[t]
\centering
\small
\caption{Quantitative results in terms of ACC and AUC on the Celeb-DFv2 dataset.}
\label{tab:celdf-finetune}
\begin{tabular}{l l l}
\hline
\textbf{Methods} & \textbf{ACC}   & \textbf{AUC}   \\ \hline\hline
F3-Net \cite{Qian2020ThinkingIF}          & 95.95\%          & 98.93\%          \\% \hline
Xception \cite{Rssler2019FaceForensicsLT}     & 97.90\%         & 99.73\%          \\ %\hline
E2E Learning \cite{cao2022end}         & \textbf{98.59\%} & \textbf{99.94\%} \\ %\hline
Ours     & 98.00\%           & 98.90\%           \\ \hline
\end{tabular}%
\end{table}

To further explore the generalization capability of our model, we follow the cross-dataset scheme presented in \cite{cao2022end}, \cite{Rssler2019FaceForensicsLT}, and \cite{Chen_Yao_Chen_Ding_Li_Ji_2021}. In this experiment, we train the model on the FaceForensics++ datasets and test its performance on the Celeb-DFv2 dataset. We present the results in Table \ref{tab:crossdataset}, where we observe that our method outperforms prior works in the area, indicating strong generalization ability in detecting deepfakes even when training is done on a different dataset and likely constitutes a different distribution (out-of-distribution).

\begin{table}[t]
\caption{Cross-dataset evaluation (AUC) by training on FaceForensics++ (LQ) and testing on the Celeb-DFv2 dataset.}
\label{tab:crossdataset}
\centering
\small
\begin{tabular}{l l}
\hline
\textbf{Methods} & \textbf{AUC} \\ \hline\hline
Xception \cite{Rssler2019FaceForensicsLT}               & 36.19\% \\ %\hline
E2E Learning  \cite{cao2022end}           & 68.71\% \\ %\hline
Local Relation Learning  \cite{Chen_Yao_Chen_Ding_Li_Ji_2021} & 78.26\% \\ %\hline
Ours                    & \textbf{80.21\%} \\ \hline
\end{tabular}%
\end{table}

To further push our approach to the limit, we explore its performance on the VLQ version of the FaceForensics++ dataset which we constructed for the first time by applying a 65\% compression ratio (see Section \ref{datasets}). We also use this dataset on two leading methods, namely DCL \cite{sun2022dual} and E2E Reconstruction Learning \cite{cao2022end}. The results are presented in Table \ref{table:robustness_65_compressed} where we observe that our method outperforms both other solutions, highlighting the efficiency and resilience of our approach in detecting deepfakes, even in the presence of highly compressed data.

To better contextualize our method within prior works, we compare the performance of our method to prior methods that have used optical flow and vision transformers in Tables \ref{table:of} and \ref{table:vit} respectively. 
% To better contextualize our method within prior works using optical flow and vision transformers, we demonstrate comparisons in Tables \ref{table:of} and \ref{table:vit} respectively. 
The results show that we achieve superior results in comparison to other methods that have used optical flow for deepfake detection when evaluating on the Face2Face manipulation method of the FaceForensics++ dataset. Similarly, we outperform other methods in the literature that leverage vision transformers.

\begin{table}[t]
\centering
\small
\caption{Quantitative results on FaceForensics++ (VLQ) dataset which is constructed by applying a 65\% compression ratio.}
\begin{tabular}{l c}
\hline
\textbf{Methods} & \textbf{ACC} \\ \hline\hline
DCL \cite{sun2022dual} & 65.20\% \\% \hline
E2E Learning \cite{cao2022end} & 78.20\% \\ %\hline
Ours & \textbf{79.70\%} \\ \hline
\end{tabular}
\label{table:robustness_65_compressed}
\end{table}

\begin{figure}[t]
  \centering
  \includegraphics[width=0.9\columnwidth]{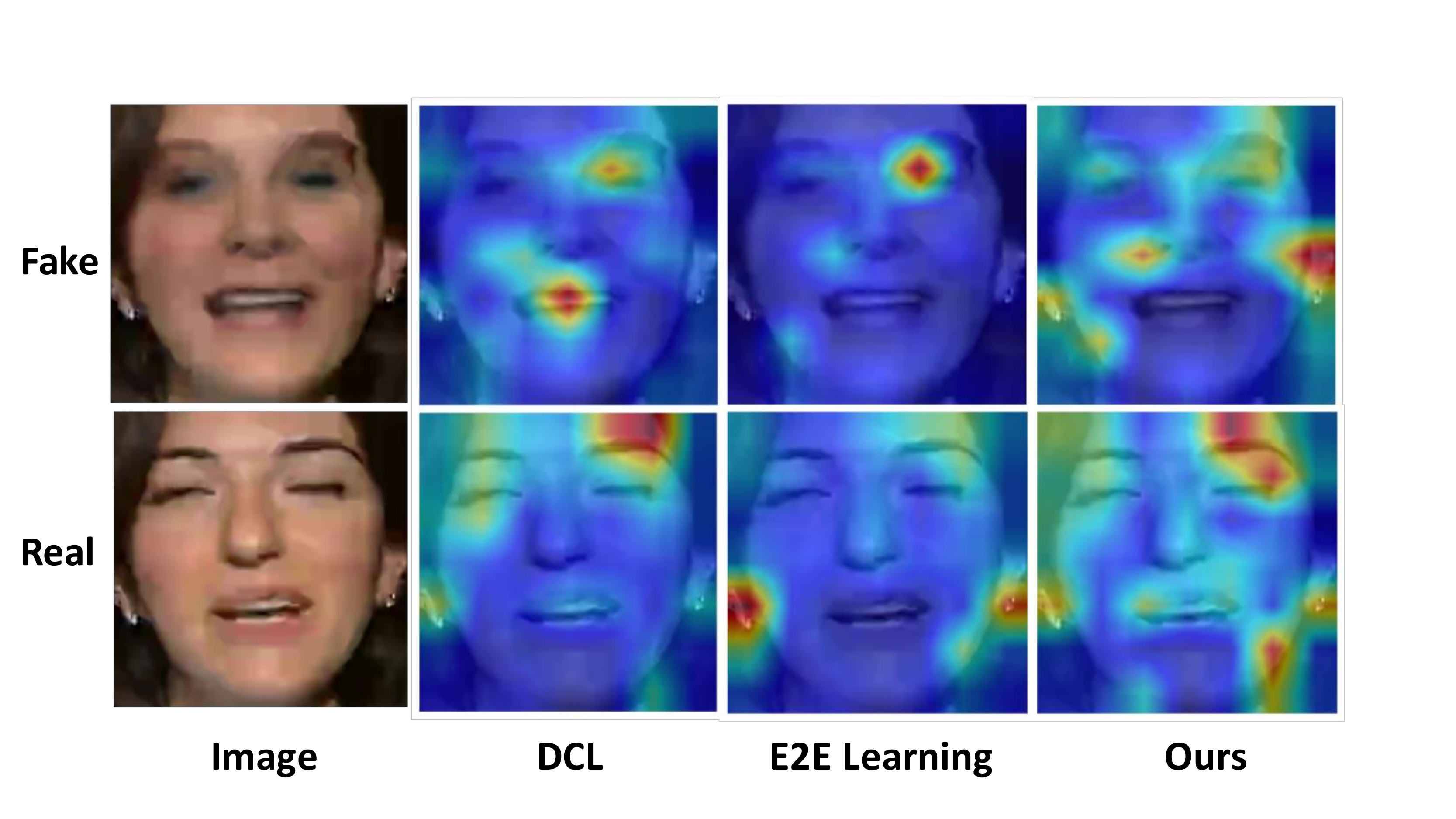}
  \caption{Comparison of Grad-CAM visualizations \cite{8237336} for our method in comparison to two other recent works.}
  \label{fig:gradcam}
\end{figure}

Lastly, we utilize Grad-CAM \cite{8237336} visualization on our model and similar performing methods to demonstrate and investigate the attention patterns of each method. Grad-CAM is capable of pinpointing the areas that the network applies more attention to, and thus deems important. We present a sample image in Figure \ref{fig:gradcam}, where the red areas highlight parts of the image which are more salient for the models. We observe that our model considers broader areas of the face image as important toward detection of whether the input is a deepfake image or not. This is a noteworthy observation as it indicates that the proposed method is capable of capturing a more comprehensive set of features and artifacts, which might be overlooked by the other models. This ability to focus on multiple areas simultaneously could enable the proposed method to better discern subtle inconsistencies and artifacts that are characteristic of deepfakes or manipulated images. In contrast, the other two models, with their more concentrated attention patterns, may be less effective in capturing the full extent of these subtle cues, which might result in lower overall performance in detecting such forgeries. Another interesting pattern which can be observed is that prior methods seem to focus on select areas, namely the left eye and to some extent the right ear. However, in addition to these regions, our method considers the nose and mouth regions, which are critical areas for authentic face images.

\begin{table}[t]
\caption{Comparison to prior deepfake detection methods that use optical flow. The results are reported on the Face2Face manipulation method of FaceForensics++.}
\small
\centering
\begin{tabular}{lll}
\hline
\textbf{Methods} & \textbf{Training dataset} & \textbf{AUC}\\ 
\hline\hline
OF + CNN \cite{Amerini2019DeepfakeVD} & FF++ & -\\ 
% \hline
OF + CNN \cite{CALDELLI202131} & FF++ & -\\ 
% \hline
OF + CNN-LSTM 
\cite{9892905} & - & 79.00 \\ 
% \hline
Ours & FF++ (LQ) & 80.21\\ 
% \hline
Ours & FF++ (HQ) & 82.19\\ 
\hline
\end{tabular}
\label{table:of}
\end{table}

\begin{table}[t]
\caption{Comparison to other ViT-based deepfake detection methods. The results are reported on FaceForensics++.}
\small
\centering
\begin{tabular}{lll}
\hline
\textbf{Methods} & \textbf{Training Dataset} & \textbf{AUC}\\ 
\hline\hline
ViXNet \cite{GANGULY2022118423} & FF++ & 74.78\\ 
% \hline
Conv ViT \cite{wodajo2021deepfake} & FF++ & 71.80\\ 
% \hline
UIA-ViT \cite{zhuang2022uiavit} & FF++ & 99.33\\ 
% \hline
Ours & FF++ (HQ) & 99.67\\ 
\hline
\end{tabular}
\label{table:vit}
\end{table}

\subsection{Ablation Studies}
In this section, we investigate the contributions of different components of our method toward facial forgery detection. As the first step, we remove the temporal consistency encoder and present the results in Tables \ref{tab:ablation-LQ}, \ref{tab:ablation-HQ}, and \ref{tab:ablation-cdf}, for FaceForensics++ (LQ), FaceForensics++ (HQ), and Celeb-DFv2, respectively. When comparing these results to the performance of our original method (also presented in each table), we observe that removing the temporal consistency encoder results in performance drops of 1.2\% to 2.9\%. This indicates the importance of learning additional temporal information through optical flow which may be difficult for the model to learn without explicit supervision.

Next, we examine the impact using simple score-level fusion in our model. To this end, we adopt two strategies instead. First, we use the joint learning approach proposed in \cite{bachmann2022multimae}, where a single pre-trained encoder accepts patches from both the RGB and optical flow modalities simultaneously. Second, instead of score-level fusion, we use feature-level fusion immediately after the embeddings are obtained from the spatial and temporal consistency encoders. The results for both experiments are presented in Tables \ref{tab:ablation-LQ}, \ref{tab:ablation-HQ}, and \ref{tab:ablation-cdf}, for the three datasets, respectively. We observe that while feature-level fusion achieves results closer to ours in comparison to joint learning, our method still obtains superior results to both these strategies.

\begin{figure}[t]
  \centering
  \includegraphics[width=0.7\columnwidth]{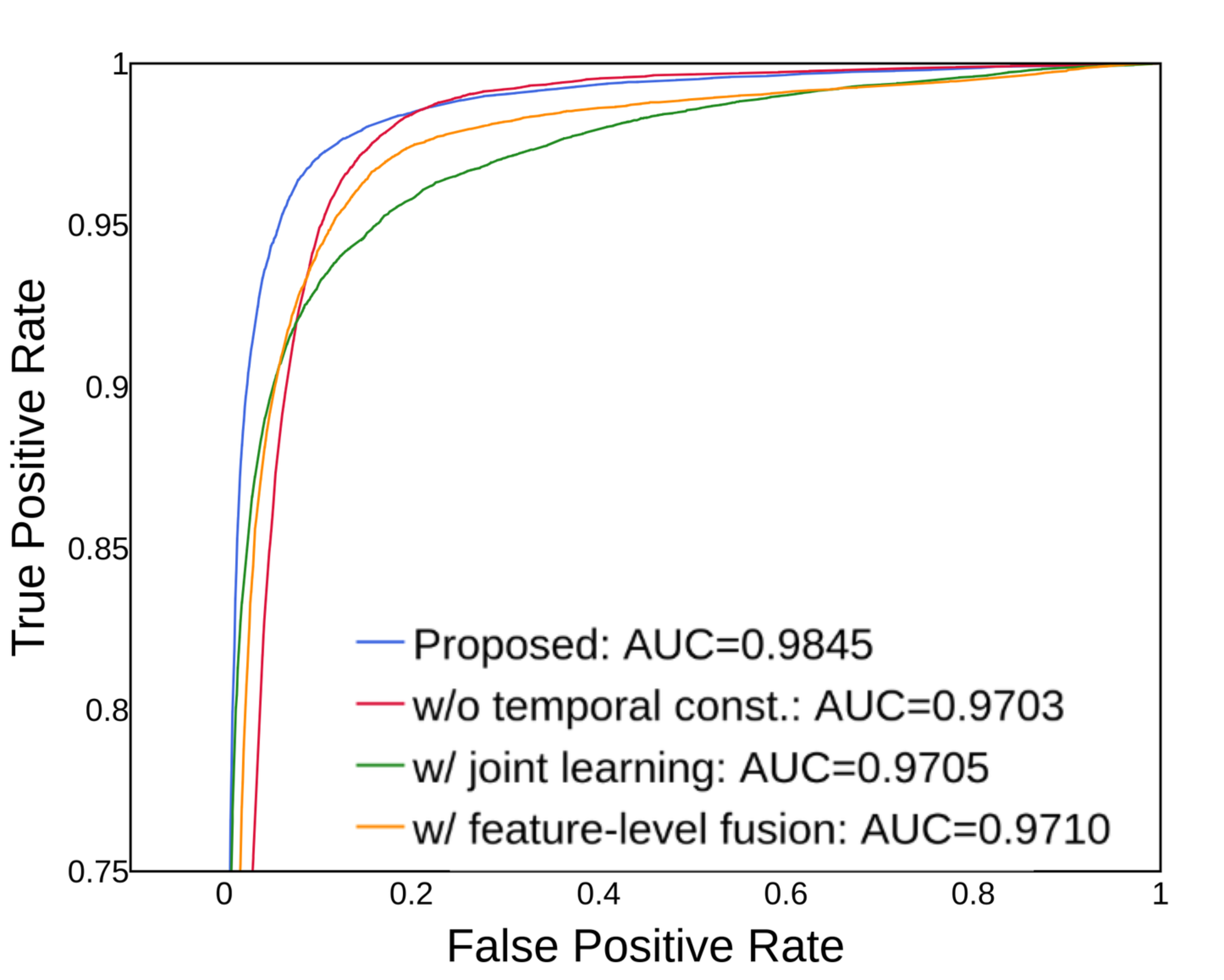}
  \caption{Receiver operating characteristic (ROC) curves for our proposed method (blue) and three ablations on the FaceForensics++ (LQ) dataset.}
  \label{fig:ablation_roc}
\end{figure}

Lastly we illustrate the Receiver Operating Characteristic (ROC) curves for our method (depicted in blue) and the three ablated variants discussed above, in Figure \ref{fig:ablation_roc}. These results are obtained on the FaceForensics++ (LQ), demonstrated in Table \ref{tab:ablation-LQ}. We observe that the true positive rates are generally higher than the model variants across different false positive rate regions, except for the version where temporal consistency is not used, which shows comparable results in true positive rates for high false positive regions. This indicates that the temporal consistency component is highly effective in reducing the number of false alarms.

%The Receiver Operating Characteristic (ROC) curves in Figure \ref{fig:ablation_roc} compare our method (in blue) with three ablated variants, using data from FaceForensics++ (LQ) in Table \ref{tab:ablation-LQ}. We observe that the true positive rates are generally higher than the model variants across different false positive rate regions, except for the version where temporal consistency is not used, which shows comparable results in true positive rates for high false positive regions. This indicates that the temporal consistency component is highly effective in reducing the number of false alarms.

\begin{table}[t]
\centering
\small
\caption{Ablation experiments on FaceForensics++ (LQ).}
% by removing the temporal consistency component, removal score-level fusion and replacing it with MultiMAE joint learning \cite{bachmann2022multimae}, and removing score-level fusion and replacing it with feature-level fusion.
% \caption{Ablation experiments on FaceForensics++ (LQ). The ablated versions include the removal of the temporal consistency component, removal of score-level fusion and replacing it with MultiMAE joint learning \cite{bachmann2022multimae}, and removal of score-level fusion and replacing it with feature-level fusion.}
\begin{tabular}{lll}
\hline
\textbf{Technique} & \textbf{ACC} & \textbf{AUC} \\
\hline\hline
Proposed & \textbf{97.79\%} & \textbf{98.45\%} \\
w/o temporal consistency & 96.51\% & 97.03\% \\
w/ joint learning \cite{bachmann2022multimae} & 95.02\% & 97.05\% \\
w/ feature-level fusion & 96.01\% & 97.10\% \\
\hline
\end{tabular}
\label{tab:ablation-LQ}
\end{table}

\begin{table}[t]
\centering
\small
\caption{Ablation experiments on FaceForensics++ (HQ).}% The ablated versions include the removal of the temporal consistency component, removal of score-level fusion and replacing it with MultiMAE joint learning \cite{bachmann2022multimae}, and removal of score-level fusion and replacing it with feature-level fusion.}
\begin{tabular}{lll}
\hline
\textbf{Technique} & \textbf{ACC} & \textbf{AUC} \\
\hline\hline
% MultiMAE \cite{bachmann2022multimae} styled joint learning 
Proposed & \textbf{98.19\%} & \textbf{99.67\%} \\
w/o temporal consistency & 96.90\% & 97.35\% \\
w/ joint learning \cite{bachmann2022multimae} & 95.81\% & 97.58\% \\
w/ feature-level fusion & 98.01\% & 99.09\% \\
\hline
\end{tabular}
\label{tab:ablation-HQ}
\end{table}

\begin{table}[t]
\centering
\small
\caption{Ablation experiments on CelebDFv2.}% The ablated versions include the removal of the temporal consistency component, removal of score-level fusion and replacing it with MultiMAE joint learning \cite{bachmann2022multimae}, and removal of score-level fusion and replacing it with feature-level fusion.}
\begin{tabular}{lll}
\hline
\textbf{Technique} & \textbf{ACC} & \textbf{AUC} \\
\hline\hline
% MultiMAE \cite{bachmann2022multimae} styled joint learning 
Proposed & \textbf{98.00\%} & \textbf{98.90\%} \\
w/o temporal consistency & 95.08\% & 97.17\% \\
w/ joint learning \cite{bachmann2022multimae} & 95.06\% & 96.55\% \\
w/ feature-level fusion & 96.81\% & 98.10\% \\
\hline
\end{tabular}
\label{tab:ablation-cdf}
\end{table}

\subsection{Limitations}
We identify several limitations in our work. First, while the integration of temporal information through optical flow improves the detection performance of our method, it also increases the computational complexity of the system, potentially limiting its real-time applicability. Second, the proposed approach may not be robust to novel deepfake techniques or attacks targeting the identified limitations. Therefore, the effectiveness and generalizability of our proposed method will need to be validated further on new datasets and deepfake scenarios as they become available in the future. Lastly, we observe that the temporal consistency contributed mostly to the reduction of false positive detection. While this can indeed be valuable for practical applications, designing additional components to further enhance the true positive detection is also of critical importance.
% Our study has limitations. The integration of optical flow, though enhancing detection, also increases computational complexity, possibly restricting real-time usage. The method's robustness against new deepfake techniques is unverified, requiring further validation on fresh datasets and scenarios. Finally, while temporal consistency reduced false positives, designing elements to boost true positive detection is equally important.

\section{Conclusion}
% In this work, 
We introduce MASDT, a learning framework for enhanced deepfake detection. Our method consists of two components, spatial and temporal consistency learning. The model follows a sequential two-step process. Initially, it employs self-supervised pre-training where both spatial learning and temporal consistency learning components engage in data reconstruction. 
Spatial learning makes use of a masked self-supervised auto-encoder to derive robust spatial features from partial facial images, while temporal consistency learning employs a similar auto-encoder to extract temporal features from partial optical flow fields. Subsequently, deepfake detection is executed through fine-tuning of the encoders of both learning components followed by simple score-level fusion. 
Various experiments on FaceForensics++ (LQ and HQ) and CelebDFv2 datasets demonstrate that our approach outperforms state-of-the-art methods by effectively learning spatial and temporal information, resulting in enhanced classification performance. 

% Several exciting avenues can be explored for future work. First, a lightweight version of our model, which could be achieved through distillation, could play a crucial role in extending the proposed method for real-time detection of facial forgeries in video streams for practical applications. Moreover, by integrating various modalities such as visual, audio, and text data and leveraging the strengths and complementary aspects of each modality, a unified framework could significantly enhance detection capabilities and overall performance through a holistic understanding of manipulated content.

% We present MASDT, a two-step deepfake detection framework. Initially, it utilizes a self-supervised pre-training strategy with spatial learning from partial facial images and temporal consistency learning from partial optical flow fields. Deepfake detection follows through fine-tuning of both encoders and score-level fusion. Tests on FaceForensics++ and CelebDFv2 datasets show our method surpasses existing ones by efficiently learning spatial and temporal data, improving classification. Future work could develop a lightweight model for real-time forgery detection and integrate multiple data modalities for a comprehensive detection framework.

\section{Acknowledgements}
This work was funded by Irdeto Canada Corporation and the Natural Sciences and Engineering Research Council of Canada (NSERC).

%%%%%%%%% REFERENCES
{\small
\bibliographystyle{ieee_fullname}
\bibliography{egbib}
}

\end{document}